\documentclass[a4paper]{article}
\usepackage{INTERSPEECH2019}
\usepackage{amsmath,graphicx,tabularx,blindtext,color,url}

\title{Lattice-based lightly-supervised acoustic model training}
\name{Joachim Fainberg, Ond\v{r}ej Klejch, Steve Renals, Peter Bell}
\address{
  Centre for Speech Technology Research, University of Edinburgh, United Kingdom}
\email{\{j.fainberg, o.klejch, s.renals, peter.bell\}@ed.ac.uk}

\begin{document}

\maketitle
\begin{abstract}
In the broadcast domain there is an abundance of related text data and partial transcriptions, such as closed captions and subtitles. This text data can be used for lightly supervised training, in which text matching the audio is selected using an existing speech recognition model. Current approaches to light supervision typically filter the data based on matching error rates between the transcriptions and biased decoding hypotheses. In contrast, semi-supervised training does not require matching text data, instead generating a hypothesis using a background language model.  
State-of-the-art semi-supervised training uses lattice-based supervision with the lattice-free MMI (LF-MMI) objective function. We propose a technique to combine inaccurate transcriptions with the lattices generated for semi-supervised training, thus preserving uncertainty in the lattice where appropriate. We demonstrate that this combined approach reduces the expected error rates over the lattices, and reduces the word error rate (WER) on a broadcast task. 

\end{abstract}
\section{Introduction}
Automatic Speech Recognition (ASR) systems are ideally trained on accurate transcripts that match the audio. There is, however, a large amount of available data that has inaccurate and partial transcripts. Examples include traditional broadcast data~\cite{bell2015mgb,driesen2013lightly,graff2002overview,long2013improving}, YouTube data~\cite{liao2013large}, medical data~\cite{mathias2005discriminative}, crowd-sourced data~\cite{van2015improving}, and children's data~\cite{nicolao2018improved}.

Obtaining manual transcriptions is expensive. Instead, there is a growing body of work on lightly supervised methods that aim to make use of partial and inaccurate transcriptions for aligning and training acoustic models. Most of these methods make use of a decode of the data made with a language model (LM) biased towards the data itself~\cite{lamel2002lightly,chan2004improving}. They typically proceed by either filtering the data at various levels of granularity~\cite{bell2015mgb,liao2013large,li2015discriminative,driesen2013lightly}, or by some combination or error correction algorithm~\cite{long2013improving,olcoz2016error,venkataraman2004efficient,chen2004lightly,van2015improving,nicolao2018improved,saz2018lightly,manohar2017jhu}. The benefit of the latter techniques is that they can often maintain more data than through filtering, while filtering is perhaps most appropriate with large amounts of data such that false rejects is a non-issue. New acoustic models can be trained on the filtered or corrected transcriptions, and sometimes the process is repeated, yielding increasingly improved models~\cite{stan2012grapheme,nicolao2018improved}.

In contrast to the lightly supervised approaches, semi-supervised training requires no transcriptions for new data, but generates hypotheses using a large (non-biased) background model. In state-of-the-art approaches using the discriminative LF-MMI objective (see~\cite{povey2016purely}), the decoding lattices are maintained and used as lattice supervision, effectively encoding the uncertainty of the hypotheses by the width of the lattice~\cite{manohar2018semi,klejch2019adapt}. As the authors remark, this is beneficial for discriminative training which is sensitive to the accuracy of the supervision~\cite{mathias2005discriminative,yu2010unsupervised}.

The contribution of this paper is three-fold. First, typical lightly supervised techniques produce single best path transcriptions on which to train. Yet, state-of-the-art semi-supervised, discriminative, training techniques benefit strongly from lattice supervision which encodes the uncertainty of the data~\cite{manohar2018semi,klejch2019adapt}. We experiment with lightly-supervised training where the lattice supervision is generated with a biased LM, and demonstrate that this can substantially improve WERs. 

Second, Long et al.~\cite{long2013improving} showed that, instead of filtering, it is possible to combine inaccurate transcripts with a biased decode lattice to create an improved best path transcription on which to train. However, the output is a best path, not a lattice. Manohar et al.~\cite{manohar2017jhu} demonstrated an algorithm that combines individual transcripts into a confusion network lattice for training, though this approach does not combine with a hypothesis lattice. We propose a new method to combine the transcriptions, and a hypothesis lattice, while, crucially, maintaining a lattice for supervision. This encodes uncertainty where the transcriptions and lattices disagree, whilst maintaining a narrow lattice where they do agree. We show up to 17.5\% relative reductions in WER with respect to a semi-supervised baseline. 

Finally, in our experiments we observed that the proposed method compensates for a large number of deletions. We propose to reduce deletions by rewarding insertions when \emph{generating the supervision lattices}, which in our experiments reduces WERs up to 13\% relative. The combined improvements from the above ideas yield up to 20\% relative WER reductions on broadcast data from the Scottish Parliament, adapting from a model trained on BBC news broadcasts.

\section{Related work}
Lamel et al.~\cite{lamel2002lightly} introduced filtering based on segment-level matching with biased LMs. This approach was later extended to discriminative training~\cite{chan2004improving}. Typically segments are filtered given a matching error rate threshold at the word (WMER) or phone level (PMER) between the transcriptions and the biased decode~\cite{bell2015mgb,lanchantin2016selection,braunschweiler2010lightly,long2013improving}. Methods operating at finer levels of granularity often include selecting \emph{islands} of consecutive words with zero string edit distance~\cite{nguyen2004light,liao2013large,driesen2013lightly,mathias2005discriminative}, or to select words based on a set of cascaded classifiers~\cite{li2015automatic}. Some approaches consider the alignment and match of two transducers, one which allows word-skips (\emph{skip-net}), and one which doesn't (\emph{sequence net})~\cite{stan2012grapheme,driesen2013lightly}, or use a factor transducer~\cite{bell15_alignment} to select reliable segments. 
Most approaches can be iterated, yielding more and better data with increasingly refined models.

Combination approaches, on the other hand, aim to maintain as much data as possible through correcting or combining the hypotheses with the transcriptions. Long et al.~\cite{long2013improving} proposed a word-level combination scheme that uses ROVER~\cite{fiscus1997post} to select a sequence of words from reference transcriptions with the corresponding hypothesis lattices with confidence scores calculated as in~\cite{seigel2011combining}. Words in the reference that occur in the lattices are given a high score to force the selection of that word. A similar approach was used by Van Dalen et al.~\cite{van2015improving} for hypotheses from crowd sourced data. Manohar et al.~\cite{manohar2017jhu} propose a different approach that combines four transcripts into a confusion network for training. In Chen et al.~\cite{chen2004lightly} the authors align the transcription to a sausage lattice version of the hypotheses (\emph{consensus network}), and select words at each arc depending on the posterior probabilites of the words and the match with the aligned word. Venkataraman et al.~\cite{venkataraman2004efficient} proposed to align the data with a robust alignment procedure, using transducers that allow for words to be skipped, and certain insertions, for which transition probabilities are estimated empirically on held-out data. The resulting best path is used as training data. A related approach was taken by Nicolao et al.~\cite{nicolao2018improved}, in which they propose to use a transducer to model variations in children's speech. In Olcoz et al.~\cite{olcoz2016error} the authors propose to correct for word-boundary errors and insertions in a lightly supervised alignment. They compare alignments and their confidences from different acoustic models, and include a model specifically trained to detect insertions. 

Most of the aforementioned techniques use a biased LM. This is typically a large background LM which is interpolated (using a high weight) towards an LM estimated on the domain of the transcriptions (e.g. \cite{driesen2013lightly}), or just the transcriptions themselves (e.g. \cite{long2013improving}).


\section{Semi-supervised LF-MMI}
The LF-MMI objective was introduced by Povey et al.~\cite{povey2016purely} as a method to train neural networks discriminatively with the MMI criterion, without requiring a first-pass of cross-entropy training. It was later extended to the semi-supervised training scenario~\cite{manohar2018semi}, and to test-time adaptation~\cite{klejch2019adapt}. We note briefly that the idea of semi-supervised LF-MMI is to generate supervision lattices by decoding unsupervised data using a seed model. Semi-supervised training, and LF-MMI, are well known, and we refer to the above papers for more details.

\section{Lattice combination}\label{sec:latcomb}
\begin{figure*}[tb]
    \centering
    \includegraphics[width=\textwidth]{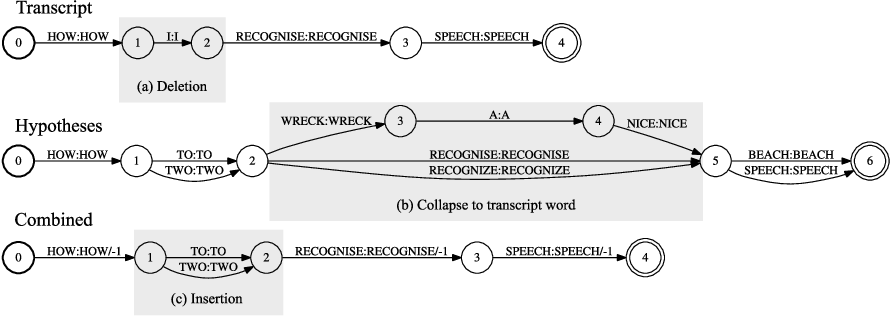}
    \caption{An example of the result of the proposed lattice combination algorithm. (a) A transcript word is deleted where it is not present in the lattice; (b) the lattice is collapsed where there is a match; (c) insertions missing in the transcriptions are kept or inserted. }
    \label{fig:lat_combination}
\end{figure*}

We present a new method to combine lattices with inaccurate transcripts. The lattices will typically be the output of a decode with a biased LM, and the transcriptions are in this paper subtitles from broadcast data. As in Long et al.~\cite{long2013improving}, we make the assumption that if a word in the transcription is present in the lattice, then the word is likely correct; but also that a transcript word not occurring in the lattices is wrong. Seen from the view of the transcriptions, then, we would like to insert, and substitute, with hypotheses from the lattice when this is the case. Viewed from the lattice, we want to collapse the lattice onto words in the transcription when possible. This provides a narrow lattice where we believe it should be confident, and wide otherwise. Consequently, if the transcriptions are not at all present in the lattice, then the lattice is kept in its entirety. An example output of the algorithm is shown in Figure~\ref{fig:lat_combination}.


To perform the combination, we require a linear transducer, $R$, of a transcription, and a hypothesis lattice $H$, for a particular utterance (we assume utterance segmentation a priori), projected on words. All weights are scaled to zero. We create an edit transducer, $E$, that allows for insertions ($\epsilon:w$), deletions ($w:\epsilon$), and substitutions ($w_i:w_j$), and compose them in the following order:
\begin{equation}
    T = R \circ E \circ H.
\end{equation}

As our goal is to maximise the number of correct words in the lattice, not to minimise the number of edits, we do not use the standard costs for $E$. Instead, we set every edit cost to $0$, apart from matches ($w_k:w_k$) which we set to $-1$. The path with the most correct words will then have the most negative total cost. We retain only the paths with that minimum cost, or some multiple of it, by pruning the transducer with a threshold $t$ times the shortest path cost:

\begin{equation}
t\otimes \left[\oplus_{\pi} w(\pi)\right],
\end{equation}
where the sum is over all paths $\pi$ in $R\circ E\circ H$. We set $t=\bar{1}$ in our experiments.

To obtain the final combined lattice, we first project the pruned transducer on the output. This retains any substitutions made by the hypothesis lattice, and deletes words in $R$ that were not present in $H$. Finally we remove epsilons, determinise, and minimise. The complete set of operations are:

\begin{equation}
    T = \mathrm{min}(\mathrm{det}(\mathrm{rmeps}(\mathrm{proj}(\mathrm{prune}(R \circ E \circ H))))).    
\end{equation}

To use the combined transducer $T$ as supervision, we add back costs by composing it with $G$, and use the resulting grammar to compile new training graphs with which we align the data to add acoustic costs.

The edit transducer $E$ is not particularly efficient in that it needs to store $(\vert V\vert+1)^2-1$ transitions for a vocabulary $V$, and the resulting search space is quadratic in the length of the input. This could be mitigated with factored transducers, three-way compositions or using a rho-matcher. 
We note, however, that the computation time spent during lattice combination is negligible compared to the decode pass required to create $H$.

Shown in Table~\ref{tab:comp_expected} are the expected WERs over the lattices with respect to the true transcriptions in the test set. The table indicates a considerable improvement with the combined lattices, and demonstrates the inaccuracy of the transcriptions.

\begin{table}[htbp]
    \caption{Expected WERs over lattices with aligned and segmented references on the test set of Scottish Parliament data.}
    \label{tab:comp_expected}
    \centering
    \begin{tabular}{l c}
    \toprule
    \textbf{Supervision} & $\mathbb{E}[\text{WER}]$ \\
    \midrule
    Transcriptions ($R$)  & $55.3$ \\
    \midrule
    Decode-biased ($H$)   & $35.5$ \\
    Combined-biased ($T$) & $26.4$ \\
    \bottomrule
    \end{tabular}
\end{table}

\section{Experimental setup}
The baseline model in this paper is trained on news data from the Multi Genre Broadcast (MGB) corpus~\cite{bell2015mgb}. The news data is filtered using MER 40\% with respect to the lightly supervised decode that is included in the MGB challenge. The resulting data consists of 179 hours across 545 shows, and approximately 15,600 (unlinked) speakers.

We use 5 hours of data from the Scottish Parliament\footnote{\url{https://www.youtube.com/user/ScottishParl}} as adaptation data. This consists of 1300 utterances across 374 speakers. On average the utterances contain 30 words each. The test set contains 6.8 hours of audio across 40 speakers. The accompanying subtitles are inaccurate, as demonstrated in Table~\ref{tab:comp_expected}. Empirically we observed large amounts of paraphrasing in the data.

\subsection{Model}
The baseline model is trained using Kaldi~\cite{povey2011kaldi}, and is based upon the 7p recipe for Switchboard. This is a factored time-delay neural network (TDNN-F) model~\cite{povey2018semi} with 12 layers, each with 1280 units (apart from the penultimate layer), and bottleneck dimensions of 256. Interleaving the layers are ReLU activations, batchnorm and dropout layers. We train on alignments obtained from a standard HMM-GMM system that matches the parameters set out in the MGB challenge~\cite{bell2015mgb}. The model is trained with speed-perturbed~\cite{ko2015audio} MGB news data for 8 epochs. The background trigram LM is trained on 640 million words of BBC subtitle text, and is restricted to the top 150,000 unigrams. We estimate biased LMs on the adaptation data in the same way, interpolating with a weight of 0.7. All models are evaluated with the background LM. During semi-supervised training, we train for 3 epochs with an initial learning rate of $5\times10^{-5}$. The lattice combination is implemented using Kaldi~\cite{povey2011kaldi} and OpenFST~\cite{allauzen2007openfst}.

\section{Experiments}
Baseline results adapting to the raw transcriptions or in a semi-supervised manner are shown in Section~\ref{sec:baseline}. Results with the lattice combination are presented in Section~\ref{sec:combination}, and with biased LMs in Section~\ref{sec:bias_lm}. We experiment with an alternative method to control for deletions in the supervision in Section~\ref{sec:deletions}.

\subsection{Baseline model}\label{sec:baseline}
The baseline model trained on BBC news data achieves $30.0$\% WER on the Scottish Parliament test data, as shown in Table~\ref{tab:baseline}. Adapting using the unfiltered transcriptions as supervision increases the error rate to $33.2$\%. This is expected given the high error rate of the transcriptions with the true reference (Table~\ref{tab:comp_expected}). The large proportion of deletion errors suggests that the transcriptions have failed to account for words present in the audio. 

In contrast, adapting in a semi-supervised manner, having generated supervision with a decode of the data, improves results. Primarily the number of deletions have dropped, which indicates that the semi-supervised supervision has filled in deletions that were absent in the transcriptions. We also note that training on the best path is worse than using the entire lattice, which is consistent with the literature~\cite{manohar2018semi,klejch2019adapt}.

\begin{table}[th]
  \caption{Baseline results on Scottish parliament data.}
  \label{tab:baseline}
  \centering
  \begin{tabular}{ l c c c c }
    \toprule
    \multicolumn{1}{l}{\textbf{Method}} & 
      \multicolumn{1}{c}{\textbf{WER \%}} &
      \textbf{Sub} & \textbf{Del} & \textbf{Ins} \\
    \midrule
    Baseline          & $30.0$ & $15.8$ & $11.3$ & $3.0$ \\
    \midrule
    Transcriptions    & $33.2$ & $11.0$ & $20.5$ & $1.7$ \\
    Semisup           & $28.6$ & $11.7$ & $14.7$ & $2.1$ \\
    Semisup-BP        & $28.8$ & $12.9$ & $13.8$ & $2.1$ \\
    \bottomrule
  \end{tabular}
\end{table}

\subsection{Lattice combination}\label{sec:combination}
Table~\ref{tab:combined+semisup} shows the results of the lattice combination method compared with purely semi-supervised approaches. The combined approach reduces WERs up to 17.5\% relative to \texttt{Semisup}, with a substantial drop in deletion and some substitution errors. As discussed above, the key difference between the combined supervision and the standard approach is that the decoded lattices typically contain multiple confusable hypotheses where they match the transcriptions, while the combined lattices will have a lattice depth of $1$ in these cases. This is reflected in the average lattice depth of the supervision lattices: $21.2$ for the combined lattices compared to $78.1$ in the original hypotheses. Additionally, the gap between best path and lattice supervision for the combined approach is larger, suggesting that it is benefiting from uncertainty encoded by the wide lattices where it was joined with the original decode.

\begin{table}[th]
  \caption{Combination and semi-supervised results.}
  \label{tab:combined+semisup}
  \centering
  \begin{tabular}{ l c c c c }
    \toprule
    \multicolumn{1}{l}{\textbf{Method}} & 
      \multicolumn{1}{c}{\textbf{WER \%}} &
      \textbf{Sub} & \textbf{Del} & \textbf{Ins} \\
    \midrule
    Baseline               & $30.0$ & $15.8$ & $11.3$ & $3.0$ \\
    \midrule
    Combined     & $23.6$ & $10.7$ & $10.7$ & $2.3$ \\
    Combined-BP  & $25.2$ & $12.2$ & $10.2$ & $2.7$ \\
    Semisup      & $28.6$ & $11.7$ & $14.7$ & $2.1$ \\
    Semisup-BP   & $28.8$ & $12.9$ & $13.8$ & $2.1$ \\
    \bottomrule
  \end{tabular}
\end{table}

\subsection{Biased language model}\label{sec:bias_lm}
The results in Table~\ref{tab:combined+semisup+bias} demonstrate the benefit of including a biased LM when generating lattice supervision. For both methods it reduces the WER by up to 10\% relative, compared to Table~\ref{tab:combined+semisup}, benefiting both lattice and best path supervision. The standard semi-supervised method seems to benefit more from biasing the LM than the combined method. In future work we would like to investigate whether this effect diminishes with a large in-domain (Scottish Parliament) LM, for which biasing will be less impactful.

\begin{table}[th]
  \caption{Results with an LM biased to the adaptation data.}
  \label{tab:combined+semisup+bias}
  \centering
  \begin{tabular}{ l c c c c }
    \toprule
    \multicolumn{1}{l}{\textbf{Method}} & 
      \multicolumn{1}{c}{\textbf{WER \%}} &
      \textbf{Sub} & \textbf{Del} & \textbf{Ins} \\
    \midrule
    Baseline               & $30.0$ & $15.8$ & $11.3$ & $3.0$ \\
    \midrule
    Combined-biased        & $23.3$ & $10.8$ & $10.2$ & $2.4$ \\
    Combined-biased-BP     & $25.0$ & $12.0$ & $10.2$ & $2.8$ \\
    Semisup-biased         & $26.8$ & $11.7$ & $13.1$ & $2.0$ \\
    Semisup-biased-BP      & $26.6$ & $12.0$ & $12.5$ & $2.1$ \\
    \bottomrule
  \end{tabular}
\end{table}


\subsection{Controlling for deletions when generating supervision}\label{sec:deletions}
The results shown thus far indicate that the seed model is inclined to delete, which has affected the generation of supervision lattices. Deletion errors account for more than half of the errors when using \texttt{Semisup} supervision. In contrast, the \texttt{Combined} method appears to control for deletions. We considered whether there is another option to compensate for a tendency to delete. We propose to achieve this by penalising deletions (or rewarding insertions) in the HCLG\footnote{HCLG denotes the composition of the following WFSTs: acoustic HMM (H), context (C), lexicon (L), and language model (G).} decoding graph when generating the lattices.  This is implemented by subtracting a constant from every word output label in the graph. We note that a standard deletion penalty is no longer current practice, as a correctly tuned system does not require it. Indeed, we did not find a penalty on the final decode to be helpful. What we are proposing is instead to include a penalty for a specific type of training, and crucially only during the generation of supervision. We decode the final model in the standard fashion.

\begin{figure}[tb]
    \centering
    \includegraphics[width=\columnwidth,trim=2 6 0 2,clip]{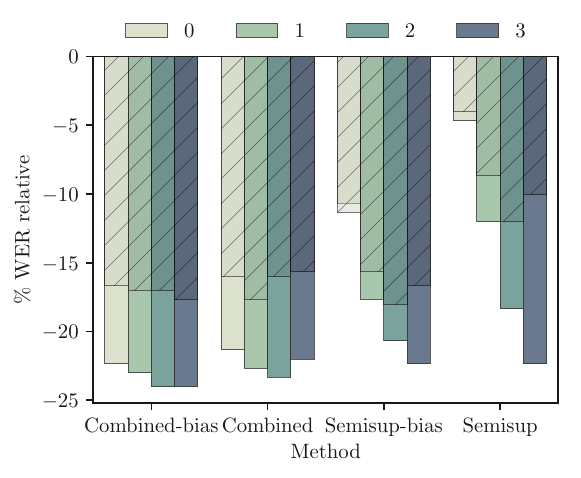}
    \caption{Results with respect to the baseline (30\%) for increasing deletion penalty / insertion reward when generating supervision
    lattices. The overlaying, hashed, bars represent the performance when using the corresponding best path supervision.}
    \label{fig:exp_lat_combination}
\end{figure}

The effect of including a deletion penalty is shown in Figure~\ref{fig:exp_lat_combination}. All models benefit strongly from the penalty. The detailed results with the optimal penalty for our experiment are shown in Table~\ref{tab:adapt_scotparl_pen-3_mgblm}. By including this penalty for supervision generation, the WERs after adaptation drop, along with the number of deletion errors, by up to 13\% relative. The combined method now yields $22.8$\% WER, a small improvement upon the previous result (Table~\ref{tab:combined+semisup+bias}), and the best result we obtained in this paper. It still improves upon standard semi-supervised training, but the difference is now less. However, tuning the deletion penalty hyperparameter is time-consuming, as it requires entire passes of decoding to generate supervision. The non-penalised combined result is close to the best overall result. 

With a deletion penalty of 3 the lattices grow very large, increasing disk usage and the time to generate training examples, by several orders of magnitude. This is reflected in the average lattice depths, which for the semi-supervised lattices is now $418.0$, compared to $10.7$ for the combined lattices. Note that the lattice depth for the combined lattices has actually reduced, since the transcriptions are likely to match to more words, as more words are present in the hypothesis lattices.

\begin{table}[th]
  \caption{Results using a deletion penalty of 3 when generating supervision lattices. Transcriptions are not affected.}
  \label{tab:adapt_scotparl_pen-3_mgblm}
  \centering
  \begin{tabular}{ l l l l l }
    \toprule
    \multicolumn{1}{l}{\textbf{Method}} & 
      \multicolumn{1}{c}{\textbf{WER \%}} &
      \textbf{Sub} & \textbf{Del} & \textbf{Ins} \\
    \midrule
    Baseline            & $30.0$ & $15.8$ & $11.3$ & $3.0$ \\
    \midrule
    Transcriptions         & $33.2$ & $11.0$ & $20.5$ & $1.7$ \\
    Combined-biased        & $22.8$ & $10.9$ & $9.0$ & $2.9$ \\
    Combined-biased-BP     & $24.7$ & $11.9$ & $9.9$ & $2.8$ \\
    Combined       & $23.4$ & $12.1$ & $7.8$ & $3.5$ \\
    Combined-BP    & $25.3$ & $12.9$ & $9.0$ & $3.4$ \\
    Semisup-biased     & $23.3$ & $11.2$ & $9.4$ & $2.7$ \\
    Semisup-biased-BP  & $25.0$ & $12.8$ & $8.4$ & $3.9$ \\
    Semisup         & $25.8$ & $12.7$ & $9.6$ & $3.5$ \\
    Semisup-BP      & $27.0$ & $13.8$ & $9.1$ & $4.1$ \\
    \bottomrule
  \end{tabular}
\end{table}

\section{Conclusions and future work}
We proposed a method for lightly supervised training, to combine inaccurate transcriptions with the decoded hypothesis lattices of a seed model. This produced an improvement upon a purely semi-supervised approach by up to $17.5$\% relative. Biasing the background language model to the data was found to substantially improve both the semi-supervised training, and the lattice combination technique. We finally suggested a deletion penalty used during the generation of the hypothesis lattices, which was effective when using a seed model that was prone to delete. The combined use of the above ideas produced a WER reduction of $20$\% with respect to the semi-supervised result.

In future work we would like to investigate improvements to the combination algorithm, how the use of stronger in- and out-of-domain language models affect the results, the effect of the pruning threshold $t$, and the above on larger amounts of data.


\section{Acknowledgements}
This work was partially supported by a PhD studentship funded by Bloomberg, and by the EU H2020 projects SUMMA (grant agreement 688139) and ELG (grant agreement 825627).

\bibliographystyle{IEEEtran}

\bibliography{bibliography}

\end{document}